%% file: main.tex
\documentclass{article}
\input{header.tex}

\title{Multiresolution Transformer Networks: Recurrence is Not Essential for Modeling Hierarchical Structure}

\author{%
  Vikas K.~Garg\thanks{Work done during an internship at Amazon.}\\
  CSAIL, MIT\\
  \texttt{vgarg@csail.mit.edu} \\
  \And
  Inderjit S.~Dhillon \\
  University of Texas at Austin \& Amazon \thanks{Work done at Amazon.} \\
  \texttt{isd@a9.com}\\
  \And
  Hsiang-Fu Yu\\
  Amazon\\
  \texttt{hsiangfu@amazon.com}
  }
  
\begin{document}
\maketitle
\input{abstract.tex}
\input{introduction.tex}
\input{setting.tex}

\input{theory.tex}

\input{mtn.tex}

\input{experiments.tex}
\input{conclusion.tex}
\clearpage
\input{bibfile.tex}
\clearpage
\input{supplementary.tex}

\end{document}

%% file: header.tex
\usepackage[preprint, nonatbib]{neurips_2019}

\usepackage{collcell}
\usepackage{booktabs}
\usepackage{amssymb, amsmath, amsthm}
\usepackage{color}
\usepackage{floatrow}
\newtheorem{proposition}{Proposition}

\usepackage[square, numbers]{natbib}
\newfloatcommand{capbtabbox}{table}[][\FBwidth]
\usepackage{blindtext}

\usepackage[utf8]{inputenc} % allow utf-8 input
\usepackage[T1]{fontenc}    % use 8-bit T1 fonts
\usepackage{url}            % simple URL typesetting
\usepackage{booktabs}       % professional-quality tables
\usepackage{amsfonts}       % blackboard math symbols
\usepackage{nicefrac}       % compact symbols for 1/2, etc.
\usepackage{microtype}      % microtypography
\usepackage{tikz}
\usepackage{siunitx}
\usetikzlibrary{shapes.geometric, arrows, calc}
\def\shadowshift{5pt,-10pt}
\def\shadowradius{10pt}
\def\aol{{\sf AOL}\xspace}
\def\amazon{{\sf OnlineX}\xspace}    

\newcommand\drawshadow[1]{
    \begin{pgfonlayer}{shadow}
        \shade[white,inner color=black,outer color=white] ($(#1.south west)+(\shadowshift)+(\shadowradius/2,\shadowradius/2)$) circle (\shadowradius);
        \shade[white,inner color=black,outer color=white] ($(#1.north west)+(\shadowshift)+(\shadowradius/2,-\shadowradius/2)$) circle (\shadowradius);
        \shade[white,inner color=black,outer color=white] ($(#1.south east)+(\shadowshift)+(-\shadowradius/2,\shadowradius/2)$) circle (\shadowradius);
        \shade[white,inner color=black,outer color=white] ($(#1.north east)+(\shadowshift)+(-\shadowradius/2,-\shadowradius/2)$) circle (\shadowradius);
        \shade[top color=black,bottom color=white] ($(#1.south west)+(\shadowshift)+(\shadowradius/2,-\shadowradius/2)$) rectangle ($(#1.south east)+(\shadowshift)+(-\shadowradius/2,\shadowradius/2)$);
        \shade[left color=black,right color=white] ($(#1.south east)+(\shadowshift)+(-\shadowradius/2,\shadowradius/2)$) rectangle ($(#1.north east)+(\shadowshift)+(\shadowradius/2,-\shadowradius/2)$);
        \shade[bottom color=black,top color=white] ($(#1.north west)+(\shadowshift)+(\shadowradius/2,-\shadowradius/2)$) rectangle ($(#1.north east)+(\shadowshift)+(-\shadowradius/2,\shadowradius/2)$);
        \shade[white,right color=black,left color=white] ($(#1.south west)+(\shadowshift)+(-\shadowradius/2,\shadowradius/2)$) rectangle ($(#1.north west)+(\shadowshift)+(\shadowradius/2,-\shadowradius/2)$);
        \filldraw ($(#1.south west)+(\shadowshift)+(\shadowradius/2,\shadowradius/2)$) rectangle ($(#1.north east)+(\shadowshift)-(\shadowradius/2,\shadowradius/2)$);
    \end{pgfonlayer}
}

\pgfdeclarelayer{shadow} 
\pgfsetlayers{shadow,main}

%% file: abstract.tex
\begin{abstract}
 The architecture of Transformer is based entirely on self-attention, and has been shown to outperform models that employ recurrence on sequence transduction tasks such as machine translation. The superior performance of Transformer has been attributed to propagating signals over shorter distances, between positions in the input and the output, compared to the recurrent architectures. We establish connections between  the dynamics in Transformer and recurrent networks to argue that several factors including gradient flow along an ensemble of multiple weakly dependent paths play a paramount role in the success of Transformer. We then leverage the dynamics to introduce {\em Multiresolution Transformer Networks} as the first architecture that exploits hierarchical structure in data via self-attention. Our models significantly outperform state-of-the-art recurrent and hierarchical recurrent models on two real-world datasets for query suggestion, namely, \aol and \amazon. In particular, on AOL data, our model registers at least 20\% improvement on each precision score, and over 25\% improvement on the BLEU score with respect to the best performing recurrent model. We thus provide strong evidence that recurrence is not essential for modeling hierarchical structure.
 
%  We thus provide strong evidence against a recent assertion that recurrence is important for modeling hierarchical structure. 
\end{abstract}

%% file: introduction.tex
\section{Introduction}
Neural methods based on recurrent or gating units \cite{HS1997, CvGBBSB2014, WY2018} have emerged as the models of choice for important sequence modeling and transduction tasks such as machine translation. These methods typically consist of an {\em encoder} that processes a stream of tokens sequentially and generates useful recurrent information that is subsequently consumed by a {\em decoder}, which produces output tokens sequentially, or as is commonly called {\em autoregressively} (though there are some exceptions, see e.g., \cite{GBXLS2018}). These methods owe their success, in large part, to their attention mechanisms that allow modeling of important dependencies in the source and target sequences by learning to focus on the most important tokens \cite{SVL2014, BCB2015, LPM2015, KDHR2017}. 
Despite their widespread success, the use of recurrent units in these models is not ideal for modeling long term dependencies due to the problem of vanishing or exploding gradients. A recent line of work mitigates this problem by stabilizing the gradient flow \cite{ZLD2018, ZLSD2018, HBFS2001}. A more radical idea, arguably, is to dispense with recurrence altogether \cite{GAGYD2017, V2017}.   

The Transformer architecture \cite{V2017} marks a recent advance that models all the dependencies between the input and the output sequences exclusively via built-in attention. This multi-layered framework, in its various incarnations, has been found to be successful across a wide range of application domains, see e.g., \cite{AKS2017, ZGMO2018, JSZK2015, DYYCLS2019,  DGVUK2019, SUV2018, GCDZ2018, PVUKSKT2018, MKMKAG2018, DCLT2018}. The success of these models is primarily ascribed to having forward and backward signals propagate over much shorter distances between the input and the output compared to  a recurrent neural net (RNN) \cite{V2017}. However, tasks such as query suggestion typically entail short input and output sequences. Therefore, it is not clear whether self-attention based models would outperform the recurrent architectures in such applications.
We formally contrast the evolution of the output in encoder and decoder of a Transformer with an RNN, and argue that a combination of several factors, including ensemble effects that are reminiscent of those underlying the success of residual nets \cite{VWB2016}, plays a key role in the success of Transformer. Note that unlike RNN based sequence models, the Transformer parallelizes a significant amount of computation at each layer. We reconcile this discrepancy in the modus operandi of these alternative notions through a novel viewpoint that postulates the RNN as a {\em masked} single layer. 
% Since the Transformer architecture consists of residual or {\em skip} connections \cite{HZRS2016, SGS2015, ZSKS2017} between its layers, our perspective helps
% % explicate the superior performance of Transformer over RNN in terms of several factors, including ensemble effects that are reminiscent of those underlying the success of residual nets \cite{VWB2016}. 

We then leverage the dynamics to design self-attention based Multiresolution Transformer Networks (MTNs) that tease out the hierarchical structure such as temporal dependencies in data.
Specifically, for applications such as query recommendation and autocompletion, contextual information as defined by a short sequence of queries becomes especially important, since the users often perform multiple search refinements in succession that reflect their search intent \cite{BHM2004, HJLHCLL2009, JKCC2014, MC2015, CJPHLCL2008}. It has been argued \cite{TBM2018} that recurrent architectures should be preferred to self-attention based networks for modeling hierarchical structure. Indeed, several hierarchical recurrent models have been proposed recently \cite{FLD2018, GBGLB2018, BLZ2018, YYDHSH2016, CAB2017}.
We contend that MTNs are natural  attention models for extracting hierarchical structure, and thus may be viewed as alternatives to the hierarchical recurrent models such as \cite{SBVLSN2015, ACW2018}.
% the first self-attention based model for  
% From a practical perspective, our analysis informs a principled plug-and-play approach to designing fully attention based modules in place of recurrent encoders and decoders. For our setting, we illustrate how an encoder and multiple decoders can be be combined functionally to yield an  encoder for what we call {\em . MTNs may be employed to  
We substantiate our assertion via strong empirical evidence that our models significantly outperform state-of-the-art (hierarchical) recurrent models on two large query datasets, namely, \amazon\footnote{Anonymized for the review period.} and \aol  \cite{PCT2006}. Moreover, we show that MTNs surpass Transformer models of similar complexity.    
 
The rest of the paper is organized as follows. We first review the Transformer architecture and the recurrent (hierarchical) sequence to sequence models in Section \ref{setting}.
We describe the dynamics in Section \ref{theory}. We then introduce MTNs in Section \ref{MTN}. The details of our experiments can be found in Section \ref{experiments}. We conclude with some future directions in Section \ref{conclusion}. To keep the exposition focused, we provide the proofs and additional experimental results in the supplementary material (Section \ref{supplementary}). %and to supplementary section \ref{supplementary}.   
 
% \section{Introduction}
 

%% file: setting.tex
\section{Background} \label{setting}
Let $X = (x_1, \ldots, x_n)$ be a sequence of token or symbol representations. Starting with an initial hidden state $h_0 = \boldsymbol{0}$ at time $t=0$, a recurrent neural net (RNN) processes symbol $x_t$, updates hidden state to $h_t$, and produces output $y_t$ at time $t \in [n] \triangleq \{1, 2, \ldots, n\}$ as 
\begin{eqnarray}
h_t & = & \phi_h(x_t W_h + h_{t-1}U_h + b_h) \label{RNN1_eq1} \nonumber \\
y_t & = & \phi_y(h_t W_y + b_y)~, 
\end{eqnarray}
where $W_h, U_h, W_y$ are weight matrices, $b_h$ and $b_y$ denote bias, $\phi_h, \phi_y$ are activation functions, and we treat $x_t$, $h_t$, and $y_t$ as row vectors.\footnote{We adopt the row based notation to improve readability (less notational clutter due to few transpose operations), and to mimic the actual flow of the standard Transformer model implementations.}
 RNNs, or alternatively, recurrent gating architectures \cite{HS1997, CvGBBSB2014, WY2018}, form the backbone of neural sequence transduction models. These models employ recurrent encoder and decoder modules, often with attention. The encoder generates a sequence of continuous representations, e.g., according to \eqref{RNN1_eq1}. The decoder then generates an output sequence of tokens one at a time using this information. Specifically, at each time the decoder takes the previously generated symbols and its current recurrent state to generate the next symbol.  During training, these models require a corpus of (source, target) pairs $(X, X')$ where the (partial) decoded sequence pertaining to $X$ is matched against the ground truth target sequence $X'$ to update the weights of the model.

The hierarchical recurrent models for query suggestion \cite{SBVLSN2015, ACW2018} strive to model the information latent in successive query reformulations during a short span. Specifically, the encoder for these models employs two levels of recurrence. The encoder treats an input {\em session} consisting of token sequences $X_1, X_2, \ldots, X_r$ that arrive in order. After a sequence $X_s, s \in [r]$ is processed at the bottom, i.e., query level, e.g., according to \eqref{RNN1_eq1}, its encoded representations are summarized (e.g. by taking their mean) and the summary is forwarded as input to the next, i.e., session level recurrent module (which in turn feeds into the decoder). Then, the next query $X_{s+1}$ in the session is processed.           

Finally, the Transformer \cite{V2017} derives inspiration from the pipeline for recurrent transduction models. However, it relies solely on attention and fully connected layers, and parallelizes the processing at each layer. Specifically, its encoder consists of a stack of $L$ layers each of which in turn consists of two sub-layers. First, position embeddings are added to input symbol representations or embeddings. The resulting representation $E_0$ is propagated up the encoder stack to get progressively refined  representations. Specifically, the bottom sub-layer at layer $\ell \in [L]$ computes multi-head attention using the embeddings emanating from the layer $\ell-1$, adds the attention to these embeddings via a skip connection, and performs layer normalization of the sum. The result is then subjected to a fully connected feed forward module, and another add and normalize step. The decoder is conceived similarly, but differs from the encoder in three important ways. First,  information leakage due to target embeddings not yet seen during training must be avoided. Second, the decoder takes the embeddings furnished by the encoder to estimate attention between source and target. Finally, the decoder produces its output representations autoregressively. At each time step, the current output representation can be treated further to estimate the probabilities for tokens. 
% as we formally establish now.  

% leverage the insights about  parallels between the Transformer and RNN

% insights from 

% The bottom sub-layer in each 

% takes the input symbol representations

% Neural sequence models  

% The Transformer   

% and $t = (t_1, \ldots, t_m)$ be the corresponding target sentence, where $s_i, i \in [n] \triangleq \{1, 2, \ldots, n\}$, are the words in the source sentence, and $t_j, j \in [m]$, in the target sentence.  We assume access to the position-encoded word embeddings $w_s = (w_{s_1}, \ldots, w_{s_n})$ and $w_t = (w_{t_1}, \ldots, w_{t_m})$, where each $w_{s_i}, w_{t_j} \in \mathbb{R}^{d}$. We assume that the source and target word embeddings are generated by the same embedding matrix $B_{emb}$. Moreover, the positions encoded in the target embeddings are assumed to be shifted right by one.  We form matrices $X_s \in \mathbb{R}^{n \times d}$ and $X_t \in \mathbb{R}^{m \times d}$ by stacking the source and target embeddings respectively. The transformer architecture consists of an encoder stack and a decoder stack, each with $L$ layers, where each layer has in-built multi-head attention and feedforward sub-layers. Rather than jumping directly into the internals of the multi-head attention, we start with a single scaled dot-product attention encoder since it already captures several main ideas of the paper. We consider multi-attention later to present a complete view.

% \noindent We try to build some mathematical intuition into the working of the Transformer architecture \cite{V2017}. We first introduce some notation. 

%% file: theory.tex
\section{Dynamics in RNN and Transformer} \label{theory}
We now draw parallels between the Transformer and the recurrent transduction models. We first introduce some notation. %We use the shorthand $[L]$ to denote the set $\{1, 2, \ldots, L\}$. 
We will often view the representations for source sequence $X = (x_1, \ldots, x_n)$ and target sequence $X' = (x_1', \ldots, x_m')$, equivalently, as matrices
$X \in \mathbb{R}^{n \times d}$ and $X' \in \mathbb{R}^{m \times d}$, where $d$ is the dimensionality of each token representation; and similarly for the accumulated outputs  $Y_t = \{y_1, \ldots, y_t\}$ of RNN, and $Y_{\ell, t} = \{y_{\ell, 1}, \ldots, y_{\ell, t}\}$ for layer $\ell$ of the Transformer. 
% Specifically, we assume that row $j$ of $X$ pertains to $x_j$, and likewise for $X'$. 
We denote activation functions by (subscripts of) $\phi$, and layer normalization by $\psi$. For the Transformer model, we assume without loss of generality that $X$ and $X'$ have already been adjusted to take positional information (including shift \cite{V2017}) into account. We denote the collection of weight matrices pertaining to multi-head attention at layer $\ell$ in Transformer by $Z_{\ell}$. All other weights in Transformer and RNN are indicated by some subscript of $W$. To simplify the notation, we will omit specifying the bias terms in our analysis (these terms can be absorbed in the weight matrices by adding an extra dimension).

While recurrence architectures often employ gating, we will focus on RNNs since they convey the essential idea underlying recurrent architectures. In contrast to the Transformer, both encoding and decoding in RNN based models proceed sequentially. So we provide a unified analysis for the evolution of  output $Y_t = \{y_1, \ldots, y_t\}$ in an RNN with time $t$, on an input representation matrix. Note that at time $t$, only partial information pertaining to first $t$ steps of the input is available to RNN. So, we mask the subsequent steps by introducing a binary matrix $M_t$ having $t$ rows and as many columns as rows in the input matrix, e.g., $M_t \in \mathbb{R}^{t \times n}$ when the input is $X \in \mathbb{R}^{n \times d}$. Specifically,  $M_t(i, i) = 1$, and $M_t(i, j) = 0$ for $j \neq i$. 
% To simplify exposition, we do not specify the functions $f_1$, $f_2$, and $f_3$. They can be readily inferred from the proofs in the Supplementary. 
Our first result describes the dynamics in RNNs. 

\begin{proposition} \label{Prop1}
The evolution of outputs of an  RNN on input $X$ can be expressed as
\begin{equation} \label{RNN}
Y_t = \phi_y(M_t \phi_h(\tilde{X} W_1) W_2)~, 
\end{equation} 
where $\tilde{X}$ depends on $X$, $W_1$, $W_2$, and encapsulates the recurrent and the input information. In particular, when the entire input is processed, the output representations are given by
\begin{equation} \label{RNN_final}
Y = \phi_y(\phi_h(\tilde{X} W_1) W_2)~. 
\end{equation}
\end{proposition}
The main intuition underlying the proof is to interpret the RNN as a fixed size layer, analogous to a decoder layer in the Transformer, that is masked in a time-dependent way to incorporate representations pertaining to only a subsequence of tokens.  We next state the evolution of output at layer $\ell$ in the encoder and the decoder of a Transformer. We need a separate treatment for the encoder and decoder stacks since the decoder operates autoregressively unlike the encoder. 
\begin{proposition} \label{Prop2}
The evolution of outputs of a Transformer encoder on input $X$ can be expressed as
\begin{equation} \label{TEncoder}
Y_{\ell} = \psi(\tilde{X}_{\ell} + \phi(\tilde{X}_{\ell} W_{\ell, 1}) W_{\ell, 2})~,  
\end{equation}
where $\ell$ is an index over layers, and $\tilde{X}_{\ell}$ depends on  $X$ and $\{Z_{r}, W_{r, 1}, W_{r, 2}: r \in [\ell]\}$.%, and $\{Z_{r}, W_{r, 1}, W_{r, 2} : r \in [\ell]\}$ denote weight matrices.
\end{proposition}

\begin{proposition} \label{Prop3}
The evolution of outputs of a Transformer decoder on its input, i.e., encoder output $Y_{enc}$ and  target representation matrix $X'$ with layer $\ell$ and time $t$ can be expressed as
\begin{equation} \label{TDecoder}
Y_{\ell, t} =  \psi\left(M_t \tilde{D}_{\ell} +  M_t \phi\left(\tilde{D}_{\ell} W_{\ell, 1} \right) W_{\ell, 2}\right)~, \end{equation}
where $\tilde{D}_{\ell}$ depends on $Y_{enc}$, $X'$, and $\{Z_r, W_{r, 1}, W_{r, 2}: r \in [\ell]\}$. In particular, when the entire input is processed, the output representations are given by
\begin{equation} \label{TDecoderFull} Y_{\ell} =  \psi\left(\tilde{D}_{\ell} +   \phi\left(\tilde{D}_{\ell}W_{\ell, 1} \right) W_{\ell, 2}\right)~.\end{equation}
\end{proposition}
We provide detailed derivations in the supplementary material. Our propositions elucidate the working of the two paradigms, i.e., attention based modeling  and recurrence based modeling. First, we are able to unravel the roles played by layers $\ell$ and time $t$ in the two philosophies. Propositions \ref{Prop1} and \ref{Prop2} make clear that the flow of information in a Transformer encoder is across the layers as opposed to RNN where the flow is across time in a single masked layer. A more important distinction is revealed about the nature of transformations encountered along the flow:  
the layers in a Transformer do not share weights and are thus less susceptible to the problem of vanishing or exploding gradients during training compared to RNN where these issues are well-known. In particular, \cite{PMB2013} argues how repeatedly applying a  transformation whose singular value falls outside a small interval leads to such problems in RNN. This robustness of a Transformer encoder is accentuated by the inclusion of $\tilde{X}_{\ell}$ via a skip connection in \eqref{TEncoder}. In particular, following the arguments of \cite{VWB2016}, it can be shown that the encoder is able to preserve the gradient flow along an ensemble of several loosely dependent short paths similar to the observed behavior in residual networks  \cite{HZRS2016, SGS2015, ZSKS2017}. Finally, a closer look at the proof of Proposition \ref{Prop2} reveals that the Transformer benefits, additionally, from computing attention between all pairs of tokens at each layer and propagating this attention to subsequent layers. In particular, viewing the input tokens as nodes of a fully connected graph, we observe that the lowest layer computes pairwise attentions between tokens directly (i.e. along the edges, or paths of length 1). The next layer assimilates attention accessible via paths of length at most 2 for any pair of tokens. The same reasoning can be extended to subsequent layers that provide progressively refined attention. 

We now compare the evolution of output in RNN \eqref{RNN} and Transformer decoder \eqref{TDecoder}. Note that a binary selection matrix appears in both the equations, which underscores the sequential processing across time. However, the decoder still benefits from the masked residual information $M_t \tilde{D}_{\ell}$, whose effect becomes pronounced with progression in $t$. 
% Note however that this benefit is alloyed due to wasted computation at multiple layers especially for small $t$ since the Transformer decoder yields refined  representations for all tokens at each $t$, albeit only a few of these representations are valid due to autoregressive decoding. This issue does not arise in the context of Transformer encoder. 
Equipped with a formal understanding of the dynamics in these models, we now introduce Multiresolution Transformer Networks (MTNs) in the context of query suggestion. Specifically, since the queries arrive one at a time, we need to mask the subsequent queries during the encoding process. Thus the corresponding part of the encoder should be similar in functionality to an RNN. In contrast, since all the tokens in a query are accessible, intra-query attention could be computed in the same way as a Transformer encoder. We describe the dynamics of 2-level MTNs, and outline how they can be extended to accommodate multiple levels of abstraction.

% clearly underscore the similarities and 

% Thus, the encoder may be treated as a multilayer architecture that takes $X_{\ell}$ as input as yields $Y_{\ell}$ as the output. Essentially, the encoder is a hierarchical network that takes the entire sentence as input, generates the self-attention in parallel, and processes it further to get a better representation with different weights at different layers (unlike RNNs where the weights are shared across time).  Note that having multiple layers helps in getting {\em indirect} or contextual attention similarities. Essentially, while the lowest layer captures pairwise attentions between words (path length 1), the next layer captures attentions between two words via paths of length at most 2 (i.e. pairwise, as well as attention via one intermediate word), and so on.  The final output $E_L$ may be viewed as a context vector that would be used for decoding.   

%% file: mtn.tex
\section{Multiresolution Transformers} \label{MTN}
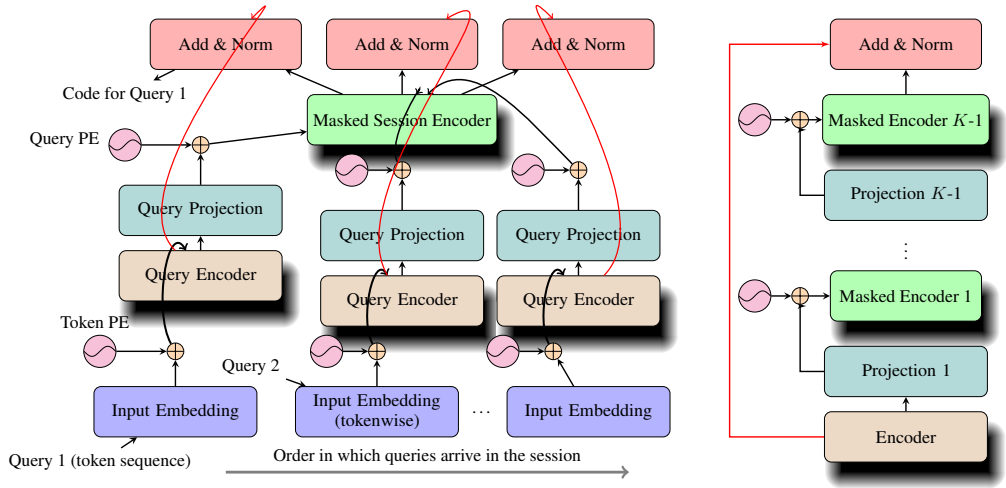
\begin{figure}[!t]
\centering
\resizebox{14cm}{!}{\input{diagram.tex}}
\caption{{\bf (Left)} Information flow through a 2-level MTN encoder for query suggestion is shown. The Query Encoder is the standard Transformer encoder, whose output token representations are subject to a projection to maintain the same model dimensionality across the levels. The Masked Session Encoder prevents information flow from subsequent queries. The shadow underneath the encoders conveys that they consist of several layers. Note that we add positional encodings (PE) for tokens in each query, and for queries in each session.  The Add \& Norm layer first computes the sum of its two arguments, one from the masked session encoder and the other via a residual or skip connection from the query encoder, and then performs layer normalization. The output codes thus obtained for each query are subsequently fed into a standard Transformer decoder (not shown in the figure) that treats the queries independently. The decoder of MTN is the same as in Transformer (i.e. single level) having the same number of layers as the Query Encoder. \textbf{(Right)}  The architecture of a general $K$-level MTN encoder is shown. Multiple masked encoders are stacked atop a standard Transformer encoder. Layers at different levels do not share any parameters.   
}\label{fig:myfigure}
\end{figure}

Despite the remarkable success of the attention based models in several domains, it is not clear whether they might be effective for tasks that possess hierarchical structure, e.g., owing to multiple temporal scales or logical composition \cite{TBM2018}. Moreover, hierarchical recurrent architectures have been shown to perform well for query suggestion \cite{SBVLSN2015, ACW2018}. Thus a natural question that arises is whether a hierarchical attention model could be designed to achieve state-of-the-art performance in such tasks. Toward that end, we introduce MTNs
that build attention at multiple levels in a principled way.

Recall that hierarchical recurrent models consist of an encoder that employs two levels of recurrence: the {\em query level} 
encodes the sequence of tokens in a query (which is within a session), and  propagates its summary to the {\em session level}, which in turn, encodes the entire sequence of queries (that form the session).
We follow the same pipeline for designing a 2-level MTN encoder by adapting the information flow between query and session levels. The query level employs a standard Transformer encoder. However, since the individual query representations arrive at the session level sequentially, we employ an a {\em Masked Session Encoder} (Fig. \ref{fig:myfigure}) that prevents  information leakage  from subsequent queries. Each layer in this encoder is similar to the middle sublayer in a Transformer decoder layer.

Note that the query level encoder for MTN generates a representation for each token in the query.  The hierarchical recurrent models typically summarize a query by taking some summary statistics such as mean of the token representations. We instead employ a linear transformation
that we achieve by what we call a {\em Query Projection} layer. Since the queries in a session may have different number of tokens, we apply zero padding to the queries before this projection.  To inform the ordering among the queries, we add positional embeddings to the individual query representations before forwarding them to the session level. The session encoder generates (masked) representations for all queries in the session. For any query $q$, we refine the individual token representations of each token $s \in q$ by adding the representation of $q$ to that of $s$ via a skip connection.  Thus the role of a session level encoder is to provide contextual information due to correlations between the queries in the session. The updated representations can then be decoded exactly as in the Transformer.  Figure \ref{fig:myfigure} shows the architecture of a 2-level MTN encoder. More generally, a $K$-level MTN architecture stacks $K-1$ masked encoders over a standard Transformer encoder. MTN consists of a single Transformer decoder. 

We now describe the evolution of the output of a 2-level MTN encoder. We add additional subscripts to differentiate between the layers of two encoders, e.g., we write $Y_{1, \ell, q}$ to denote the embeddings for query $q$ produced by layer $\ell$ of query level (i.e., level 1) encoder.
Likewise, $W_{2, \ell, 1}, W_{2, \ell, 2}$ denotes the weights for layer $\ell$ of session encoder. We denote the  weights of the query projection layer by $w_{proj}$. Our next result elucidates how MTN builds on benefits, e.g., ensemble effects and progressively refined attention, inherited from the Transformer, by leveraging important multiresolution information.  
% , and the embeddings output by MTN encoder for all queries until query $q$ by $\tilde{Y}_{2, L_2, t(q)}$
%The following equations follow immediately from Proposition \ref{Prop2} and Proposition \ref{Prop3}.  
% \begin{eqnarray*}
% \tilde{X}_{\ell} & = & f_2(X, \{Z_{1, r}, W_{1, r, 1}, W_{1, r, 2}: r \in [\ell]\}), \hspace{3cm} \ell \in [L_1]\\
% Y_{1, \ell} & = & \psi(\tilde{X}_{\ell} + \phi_1(\tilde{X}_{\ell} W_{1, \ell, 1}) W_{1, \ell, 2}), \hspace{4.1cm} \ell \in [L_1] \\
% \tilde{D}_{2, \ell} & = & f_3(Y_{1, L_1}, Y_{1, L_1}W_{proj}, \{Z_{2, r}, W_{2, r, 1}, W_{2, r, 2}: r \in [\ell]\}) \hspace{0.8cm} \ell \in [L_2] \\
% Y_{2, \ell, t} & = & \psi\left(M_t \tilde{D}_{2, \ell} +  M_t \phi\left(\tilde{D}_{2, \ell} W_{2,\ell, 1} \right) W_{2, \ell, 2}\right) \hspace{2.5cm} \ell \in [L_2]\\
% Y_t & = & \psi(Y_{2, L_2, t} + Y_{1, L_1})
% \end{eqnarray*}
\begin{proposition} \label{Prop4}
Let $n$ be the maximum number of tokens in any query. The evolution of outputs of a 2-level MTN encoder, having $L_1$ layers in the query level and $L_2$ layers in the session level, on a session $Q_S = \{q_{1}, \ldots, q_{|S|}\}$ with  query $q \in Q_S$ at position $t(q) \in [|S|]$ can be expressed as 
\begin{eqnarray} \label{MTNEncoder}
\tilde{Y}_{2, L_2, t(q)} & = & \psi\left(M_{n t(q)}(Y_{1, L_1} ~+~ \hat{Y}_{2, L_2})\right)~, ~~~ \text{  where  }\\
Y_{1, L_1, q} & = & \psi \left(\tilde{X}_{q, L_1}  +  \phi \left(\tilde{X}_{q, L_1} W_{1, L_1, 1} \right) W_{1, L_1, 2}\right)~,~~~~~ \text{ and }\\
\hat{Y}_{2, L_2, q} & = & C_{t(q)} \psi\left(f(Y_{1, L_1}) +  \phi\left(f(Y_{1, L_1}) W_{2, L_2, 1} \right) W_{2, L_2, 2}\right)~,
\end{eqnarray}
where $\tilde{Y}_{2, L_2, t(q)}$ denotes output embeddings for $q$ and the queries preceding $q$; $M_{n t(q)} \in \mathbb{R}^{n t(q) \times n|S|}$; $\tilde{X}_{q, L_1}$ depends on input embeddings $X_q$ of query $q$, attention weights $Z_{1, r}, r \in [L_1]$, and weights $W_{1, \ell, 1}, W_{1, \ell, 2}$ for $\ell \in [L_1-1]$; $C_{t(q)} \in \mathbb{R}^{n \times |S|}$ contains 1 at each entry in column $t(q)$, and 0 everywhere else; and $f$ is parameterized with  $Z_{2, r}, r \in [L_2]$, and $W_{2, \ell, 1}, W_{2, \ell, 2}$ for $\ell \in [L_2-1]$~.  
%  $\ell$ is an index over layers, and $\tilde{X}_{\ell} = f_2(X, \{Z_{r}, W_{r, 1}, W_{r, 2}: r \in [\ell]\})$ for some function $f_2$.
\end{proposition}
The dynamics in MTN decoder are  similar to Proposition \ref{Prop3} except that $Y_{enc}$ is replaced by the output of the MTN encoder. We now provide strong empirical evidence to substantiate the efficacy of MTN.         
% \begin{equation} 
% Y_{\ell, t} =  \psi\left(M_t \tilde{D}_{\ell} +  M_t \phi\left(\tilde{D}_{\ell} W_{\ell, 1} \right) W_{\ell, 2}\right)~, \end{equation}

% \begin{equation} \label{MTNEncoder}
% Y_{k, \ell, t} = 

% \psi(\tilde{X}_{\ell} + \phi_1(\tilde{X}_{\ell} W_{\ell, 1}) W_{\ell, 2})~,  
% \end{equation}

% \item Our approach presents a plug-and-play approach where the recurrent components can be replaced by an attention component. For instance, as a consequence, we can simply write the following equations for the evolution of a multiresolution encoder with $L_k$ layers at level $k \in [K]$.

% \color{cyan} 
% \begin{itemize}

% \item 

% \begin{equation} 
% Y_{\ell, t} =  \psi\left(M_t \tilde{D}_{\ell} +  M_t \phi\left(\tilde{D}_{\ell} W_{\ell, 1} \right) W_{\ell, 2}\right)~, \end{equation}

% \begin{equation} \label{MTNEncoder}
% Y_{k, \ell, t} = 

% \psi(\tilde{X}_{\ell} + \phi_1(\tilde{X}_{\ell} W_{\ell, 1}) W_{\ell, 2})~,  
% \end{equation}

% \end{itemize}
% \color{black}

%% file: diagram.tex
% \tikzstyle{AddNorm} = [rectangle, rounded corners, minimum width=3cm, minimum height=1cm,text centered, draw=black, fill=red!30]
% \tikzstyle{io} = [trapezium, trapezium left angle=70, trapezium right angle=110, minimum width=3cm, minimum height=1cm, text centered, draw=black, fill=blue!30]
% \tikzstyle{process} = [rectangle, minimum width=3cm, minimum height=1cm, text centered, text width=3cm, draw=black, fill=orange!30]
% \tikzstyle{decision} = [diamond, minimum width=3cm, minimum height=1cm, text centered, draw=black, fill=green!30]
% \tikzstyle{arrow} = [thick,->,>=stealth]

% some parameters for customization

\tikzstyle{SEnc} =  [fill=green!30,rectangle,rounded corners,minimum height=2cm,minimum width=2cm] 
%\begin{tikzpicture}[node distance=2cm]
\tikzstyle{AddNorm} = [rectangle, rounded corners, minimum width=3cm, minimum height=1cm,text centered, draw=black, fill=red!30]
\tikzstyle{SEnc} = [rectangle, rounded corners, minimum width=3cm, minimum height=1cm,text centered, draw=black, fill=green!30]
\tikzstyle{io} = [trapezium, trapezium left angle=70, trapezium right angle=110, minimum width=3cm, minimum height=1cm, text centered, draw=black, fill=blue!30]
\tikzstyle{linear} = [rectangle, rounded corners, minimum width=3cm, minimum height=1cm, text centered, text width=3cm, draw=black, fill=teal!30]
\tikzstyle{decision} = [diamond, minimum width=3cm, minimum height=1cm, text centered, draw=black, fill=green!30]
\tikzstyle{arrow} = [thick,->,>=stealth]
\tikzstyle{Empty} =  [rectangle,rounded corners,minimum height=0cm,minimum width=0cm]
\tikzstyle{line} = [draw, -latex', thick, color=red]

\tikzstyle{dottedarrow} = [thick,->,>=stealth]

\begin{tikzpicture}[node distance=1.5cm, plus/.style={path picture={
  \draw[black]
(path picture bounding box.south) -- (path picture bounding box.north) (path picture bounding box.west) -- (path picture bounding box.east);
}}]
%\centering
\node (AN1) [AddNorm] {Add \& Norm};
\node (AN2) [AddNorm, right of=AN1, xshift=2cm] {Add \& Norm};
\node (AN3) [AddNorm, right of=AN2, xshift=2cm] {Add \& Norm};
\node (AN1K) [AddNorm, right of=AN3, xshift=5cm] {Add \& Norm};
\node (in1K) [SEnc, below of=AN1K] {Masked  Encoder $K$-1};
%\node at ($(E1)!.5!(E3)$) {\ldots};
\node (L1K) [linear, below of=in1K] {Projection $K$-1};
%\node (AN11) [AddNorm, below of=L1K] {Add \& Norm};
\node (in11) [SEnc, below of=L1K, yshift=-0.5cm] {Masked  Encoder 1};
\node at ($(L1K)!.5!(in11)$) {\vdots};
\node(Empty11) [Empty, left of=AN1K, xshift=-2cm]{}; 
\node (L11) [linear, below of=in11] {Projection 1};
\node (Q11) [linear, below of=L11, fill=brown!30, yshift=0.2cm] {Encoder};
\path [line] (Q11) --(Q11-|Empty11) |- (AN1K);
%\path [line] (Q11) --(Q11-|Empty11) |- (AN1K);
%\draw[->, line width=0.3mm, color=red] (Q11)  to [out=135,in=45] (in11);
\node (in1) [SEnc, below of=AN2] {Masked Session Encoder};
\drawshadow{in1}
\draw [arrow] (in1) -- (AN1);
\draw [arrow] (in1) -- (AN2);
\draw [arrow] (in1) -- (AN3);

\draw [arrow] (in1K) -- (AN1K);
%\draw [arrow] (L1K) -- (in1K);
%\draw [arrow] (L11) -- (in11);
\draw [arrow] (Q11) -- (L11);
\drawshadow{in1K};
\drawshadow{in11};
\drawshadow{Q11};

\node (op1K) [draw,circle, fill=magenta!30, inner sep=-0.4pt, left of=in1K, xshift=-1.5cm] 
{\tikz \draw[scale=0.10,domain=-3.141:3.141,smooth,variable=\t]
plot (\t,{sin(\t r)});};
\node (op11) [draw,circle, fill=magenta!30, inner sep=-0.4pt, left of=in11, xshift=-1.5cm] 
{\tikz \draw[scale=0.10,domain=-3.141:3.141,smooth,variable=\t]
plot (\t,{sin(\t r)});};
\node (p1K) [draw,circle,plus,minimum width=0.25 cm, fill=orange!30, right of=op1K, xshift=-0.6cm]{}; 
\node (p11) [draw,circle,plus,minimum width=0.25 cm, fill=orange!30, right of=op11, xshift=-0.6cm]{}; 
\draw [arrow] (p1K) -- (in1K);
\draw [arrow] (p11) -- (in11);
\draw [arrow] (op1K) -- (p1K);
\draw [arrow] (op11) -- (p11);
%\path [line, color=black] (L1K) --(L1K-|p1K) |- (p1K);
\node(Empty1K) [Empty, below of=p1K, yshift=1.3cm]{}; 
\path [line, color=black] (L1K) -- (L1K-|Empty1K) |- (Empty1K);
\node(Empty112) [Empty, below of=p11, yshift=1.3cm]{}; 
\path [line, color=black] (L11) -- (L11-|Empty112) |- (Empty112);

\node (op1) [draw,circle, fill=magenta!30, inner sep=-0.4pt, below of=in1, yshift=0.5cm, xshift=-1cm] 
{\tikz \draw[scale=0.10,domain=-3.141:3.141,smooth,variable=\t]
plot (\t,{sin(\t r)});};
\node (op2) [draw,circle, fill=magenta!30, inner sep=-0.4pt, left of=op1, xshift=-3cm, yshift=0.5cm, label={[xshift=-1.2cm, yshift=-0.5cm, color=black]Query PE}] 
{\tikz \draw[scale=0.10,domain=-3.141:3.141,smooth,variable=\t]
plot (\t,{sin(\t r)});};
\node (op3) [draw,circle,fill=magenta!30, inner sep=-0.4pt, right of=op1, xshift=2cm] 
{\tikz \draw[scale=0.10,domain=-3.141:3.141,smooth,variable=\t]
plot (\t,{sin(\t r)});};

\node (p1) [draw,circle,plus,minimum width=0.25 cm, fill=orange!30, right of=op1, xshift=-0.5cm]{}; 
\node (p2) [draw,circle,plus,minimum width=0.25 cm, fill=orange!30, right of=op2]{};
\node (p3) [draw,circle,plus,minimum width=0.25 cm, fill=orange!30, right of=op3, xshift=-0.5cm]{};

%\draw [arrow] (p1) -- (in1);
\draw [arrow] (p2) -- (in1);
% \draw [arrow] (p3) -- (in1);
\draw [arrow] (op1) -- (p1);
\draw [arrow] (op2) -- (p2);
\draw [arrow] (op3) -- (p3);
\draw[->, line width=0.3mm] (p3)  to [out=135,in=45] (in1);
\draw[->, line width=0.3mm] (p1)  to [out=120,in=60] (in1);

\node (L1) [linear, below of=p1, yshift=0.2cm] {Query Projection};
\node (L2) [linear, below of=p2, yshift=0.2cm] {Query Projection};
\node (L3) [linear, below of=p3, yshift=0.2cm] {Query Projection};

\draw [arrow] (L1) -- (p1);
\draw [arrow] (L2) -- (p2);
\draw [arrow] (L3) -- (p3);

\node (Q1) [linear, below of=L1, fill=brown!30, yshift=0.2cm] {Query Encoder};
\drawshadow{Q1}
\node (Q2) [linear, below of=L2, fill=brown!30,  yshift=0.2cm] {Query Encoder};
\drawshadow{Q2}
\node (Q3) [linear, below of=L3, fill=brown!30, yshift=0.2cm] {Query Encoder};
\drawshadow{Q3}
\draw [arrow] (Q1) -- (L1);
\draw [arrow] (Q2) -- (L2);
\draw [arrow] (Q3) -- (L3);

\node (rp1) [draw,circle, fill=magenta!30, inner sep=-0.4pt, below of=Q1, yshift=0.5cm, xshift=-1.5cm] 
{\tikz \draw[scale=0.10,domain=-3.141:3.141,smooth,variable=\t]
plot (\t,{sin(\t r)});};
\node (rp2) [draw,circle, fill=magenta!30, inner sep=-0.4pt, left of=rp1, xshift=-3cm, label={[xshift=-0.1cm, color=black]Token PE}] 
{\tikz \draw[scale=0.10,domain=-3.141:3.141,smooth,variable=\t]
plot (\t,{sin(\t r)});};
\node (rp3) [draw,circle,fill=magenta!30, inner sep=-0.4pt, right of=rp1, xshift=2cm] 
{\tikz \draw[scale=0.10,domain=-3.141:3.141,smooth,variable=\t]
plot (\t,{sin(\t r)});};

\node (tp1) [draw,circle,plus,minimum width=0.25 cm, fill=orange!30, right of=rp1, xshift=-0.5cm]{}; 
\node (tp2) [draw,circle,plus,minimum width=0.25 cm, fill=orange!30, right of=rp2]{};
\node (tp3) [draw,circle,plus,minimum width=0.25 cm, fill=orange!30, right of=rp3, xshift=-0.5cm]{};

%\draw [arrow] (tp1) -- (Q1);
%\draw [arrow] (tp2) -- (Q2);
%\draw [arrow] (tp3) -- (Q3);
\draw [arrow] (rp1) -- (tp1);
\draw [arrow] (rp2) -- (tp2);
\draw [arrow] (rp3) -- (tp3);

\draw[->, line width=0.4mm] (tp1)  to [out=120,in=120] (Q1);
\draw[->, line width=0.4mm] (tp2)  to [out=120,in=120] (Q2);
\draw[->, line width=0.4mm] (tp3)  to [out=120,in=120] (Q3);

\node (E1) [linear, below of=tp1,fill=blue!30, yshift=0.3cm] {Input Embedding (tokenwise)};
\node (E2) [linear, below of=tp2,fill=blue!30, yshift=0.3cm] {Input Embedding};
\node (E3) [linear, below of=tp3,fill=blue!30, yshift=0.3cm, xshift=0.7cm] {Input Embedding};

\draw [arrow] (E1) -- (tp1);
\draw [arrow] (E2) -- (tp2);
\draw [arrow] (E3) -- (tp3);
\draw [ultra thick, gray, ->] (0,-8.5) -- ++(8.0cm,0) node [above, pos=0.5, color=black] {Order in which queries arrive in the session};

\draw[->, line width=0.25mm, red] (Q1)  to [out=120,in=30] (AN2);
\draw[->, line width=0.25mm, red] (Q2)  to [out=135,in=45] (AN1);
\draw[->, line width=0.25mm, red] (Q3)  to [out=45,in=135] (AN3);
\node (emp2) [Empty, left of=E2, yshift=-1cm] {Query 1 (token sequence)};
\draw [arrow] (emp2) -- (E2);

\node (emp1) [Empty, left of=E1,yshift=0.9cm,xshift=-1cm] {Query 2};
\draw [arrow] (emp1) -- (E1);

% \node (emp1) [Empty, below of=E1] {Session Query 2};
% \draw [arrow] (emp1) -- (E1);

\node at ($(E1)!.5!(E3)$) {\ldots};

\node (out1) [Empty, left of=AN1, xshift=-0.5cm, yshift=-1cm] {Code for Query 1};
\draw [arrow] (AN1) -- (out1);
\end{tikzpicture}

% \node (out2) [Empty, above of=AN2] {Output 1};
% \draw [arrow] (AN2) -- (out2);

% \node (emp1) [Empty, below of=E1] {Query 2};
% \draw [arrow] (emp1) -- (E1);

% \draw[->, line width=0.25mm, olive] (in1)  to [out=135,in=45] (in1);

%\draw [arrow] (Q2) -- (AN1);

% \node (op2) [draw,circle,inner sep=-0.4pt] at (0,0)
% {\tikz \draw[scale=0.15,domain=-3.141:3.141,smooth,variable=\t]
% plot (\t,{sin(\t r)});};
% \node (op3) [draw,circle,inner sep=-0.4pt] at (0,0)
% {\tikz \draw[scale=0.15,domain=-3.141:3.141,smooth,variable=\t]
%plot (\t,{sin(\t r)});};

% \node (pro1) [process, below of=op1] {Process 1};
% \node (dec1) [decision, below of=pro1, yshift=-0.5cm] {Decision 1};
% \node (pro2a) [process, below of=dec1, yshift=-0.5cm] {Process 2a text text text text text text text text text text};
% \node (pro2b) [process, right of=dec1, xshift=2cm] {Process 2b};
% \node (out1) [io, below of=pro2a] {Output};
% \node (stop) [AddNorm, below of=out1] {Stop};

% \draw [arrow] (AN1) -- (in1);
% \draw [arrow] (in1) -- (pro1);
% \draw [arrow] (pro1) -- (dec1);
% \draw [arrow] (dec1) -- node[anchor=east] {yes} (pro2a);
% \draw [arrow] (dec1) -- node[anchor=south] {no} (pro2b);
% \draw [arrow] (pro2b) |- (pro1);
% \draw [arrow] (pro2a) -- (out1);
% \draw [arrow] (out1) -- (stop);

%% file: experiments.tex
\begin{table}
  \caption{Details of Datasets}
  \label{table:data}
  \centering
\begin{tabular}{crrrr}
\toprule
 &\multicolumn{2}{c}{\bf \amazon}
&
\multicolumn{2}{c}{\bf \aol} 
\\\cmidrule(r){2-3} \cmidrule(l){4-5}
 & {\it Sessions} & {\it Unrolled Query Pairs}  & {\it Sessions} & {\it Unrolled Query Pairs}      \\
\midrule
{\bf Training} & $4,092,254$ & $10,964,531$ & $1,628,433$ & $4,285,507$ \\
{\bf Validation} & $84,111$ & $225,775$ & $90,932$ & $238,890$ \\
{\bf Test} & $42,349$ & $113,579$ & $90,507$ & $238,289$\\
\bottomrule
\end{tabular}
\end{table}

\section{Experiments} \label{experiments}
We demonstrate the merits of our approach via a detailed analysis of our experiments on two search logs, namely, \aol and \amazon. The objective of our experiments is two-fold. First, \cite{TBM2018} suggested that fully attentional models such as Transformer are not suitable for modeling hierarchical structure in natural language processing tasks, and recurrent architectures perform substantially better. We provide strong empirical evidence that our MTNs, despite relying entirely on attention, significantly outperform state-of-the-art  (hierarchical) recurrent models on both these datasets. Second, our results elucidate that modeling the multiresolution structure is indeed important. Specifically, for Transformer and MTN models of comparable complexity in terms of number of parameters and total number of layers, having session layers bestows MTN models with considerably better performance than the Transformer models. We first describe the two datasets and the experimental setup. %We now describe the evaluation criteria and the results of our experiments.
% It has been argued recently \cite{TBM2018} that transformer networks are not suitable for modeling hierarchical structure in natural language processing tasks, and recurrent architectures should be preferred.  We contend that 
% MTNs are the true attention counterparts to the hierarchical recurrent models such as \cite{SBVLSN2015, ACW2018}. We demonstrate strong empirical evidence that our multiresolution models significantly outperform state-of-the-art sequence to sequence and hierarchical recurrent models on two large datasets, namely, \amazon Search Log data and \aol search log data \cite{PCT2006}.  

% such as whether MTNs could 

% such as (a) whether

% MTNs could outperform the other 
\subsection{Description of datasets}
The \aol data \cite{PCT2006} consists of 16,946,938 queries (and their timestamp) submitted by 657,426 unique anonymous users between March 1, 2006 and May 31, 2006.  We used the evaluation approach suggested in \cite{BSBK2007}. Specifically, we assume a new session whenever no queries were issued for at least 30 minutes, and filter the sessions based on their lengths (minimum 3 and maximum 5). As suggested in \cite{ACW2018}, we removed all successive duplicate queries from each session, considered queries with length at most 10, and randomly partitioned the sessions into training (95\%), validation (2.5\%), and test (2.5\%) sessions. We obtained our data by treating every pair of consecutive queries as a source session-target query in the same way as \cite{ACW2018}. Thus, for a session of length $k$, we can construct $k-1$ such source-session target-query pairs.
% Thus, each session with $k$ queries contributed $k-1$ query pairs when unrolled, since each query with the exception of  first and last queries in a session was included in two pairs, as source in one and as target in other. 
We constructed a vocabulary from training data by including all words with at least 8 occurrences, and replaced all the other words by an <unk> token.  This resulted in a vocabulary of 76,604 unique tokens. We also collected logs from \amazon for a period of two months in 2018. In particular, for each day, we randomly sampled a small amount of unique sessions. Specifically, 55/4/2 days of the sampled sessions were used for training, validating, and testing, respectively. %The \amazon dataset, similarly, was compiled from searches issued by \amazon customers over a 2 month period in 2018. We randomly sampled from 55 days to create training set, 4 days to create validation set, and remaining 2 days to create test set.
We applied the same pre-processing steps as for \aol, and obtained a vocabulary of 81,893 unique tokens. Table \ref{table:data} shows the statistics of the data for our experiments. 

\subsection{Experimental setup}
We compared our method to four state-of-the-art transduction models, namely, Seq2Seq with global attention \cite{LPM2015}, Hierarchical LSTM (H-LSTM) \cite{SBVLSN2015}, M-NSRF \cite{ACW2018}, and Transformer \cite{V2017}. Among these, M-NSRF was proposed to jointly perform document ranking and query suggestion. Since we do not consider the task of document ranking, we discarded the ranking component of the architecture. The resulting architecture is similar to H-LSTM with one major exception, namely, the former suggests using an entropy regularization term in the cross entropy loss function to prevent distribution over output tokens from being too skewed.  
We learned 300-dimensional word embeddings from scratch (i.e. without using pretrained word2vec or glove embeddings), and set the dropout rate to 0.1 in each case \cite{SHKSS2014}. The model output dimension was set to 512, and the dimension of projection layers to 1024, for both the Transformer and our method (as suggested in \cite{V2017}). Likewise, the other methods employed bidirectional LSTMs where each direction yielded a 256-dimensional vector, thereby resulting in a 512 dimensional recurrent state vector.   
The batch size was chosen in each case to accommodate as much data as possible subject to ensuring training could be accomplished with a single GPU memory. Moreover, for the methods with the session level encoder (H-LSTM, M-NSRF, and MTN), we formed batches by grouping sessions based on the number of queries they contained, so that maximum data could be accommodated in each batch. For each baseline, we performed model selection by training the corresponding architecture  for 5 epochs and choosing the model with the least validation error. We found that the different methods required approximately the following wall clock time per epoch: Seq2Seq and H-LSTM (2 hours), M-NSRF and
Transformer (2.5 hours), and MTN (3 hours). We employed multi-head attention with 8 heads for both the Transformer and our method, and followed the same optimization schedule, including 4000 warm up steps, as suggested in \cite{V2017}. We experimented with label smoothing \cite{V2017} for both Transformer and MTN models. We found that MTN model achieved best level of performance with smoothing $0.05$ after 2 epochs, or $0.01$ after  5 epochs. Transformer performed well however with little to no smoothing.  We used a dropout rate 0.1 \cite{SHKSS2014} for all the models. We found empirically that M-NSRF performed best when the hyperparameter pertaining to entropy regularization was set to 0.1 \cite{ACW2018} and learning rate to 0.001. Likewise, we optimized the hyperparameters for all other models based on their validation error. All our models were implemented in PyTorch and executed on a single GPU.  

\begin{table}
  \caption{$n$-gram precision scores for the different models on the two datasets}
  \label{table:ngram}
  \centering
\begin{tabular}{ccccc}
\toprule
 &\multicolumn{2}{c}{\bf \amazon}
&
\multicolumn{2}{c}{\bf \aol} 
\\\cmidrule(r){2-3} \cmidrule(l){4-5}
 & {{\it Size} (MB)} & {\it 1/2/3/4-gram}  & {{\it Size} (MB)} & {\it 1/2/3/4-gram}      \\
\midrule
{\bf Seq2Seq Attn.} & $449$ & $35.9/22.1/13.6/9.1$ & $421$ &  $28.9/13.9/9.7/8.3$ \\
{\bf H-LSTM} & $503$ & $35.4/20.2/12.3/8.1$ & $475$ & $26.5/10.6/6.3/4.5$ \\
{\bf M-NSRF} & $503$ & ${\bf 36.0}/20.7/12.9/8.6$ & $475$ & $27.0/11.0/6.7/4.9$\\
{\bf Transformer} & $462$ & $34.0/20.6/12.6/7.6$  &   $419$  &$34.2/19.6/13.1/8.9$ \\
{\bf MTN (Ours)}  & $465$ & $35.1/{\bf 26.4}/{\bf 19.0}/{\bf 13.1}$ & $407$  & ${\bf 35.7}/{\bf 21.5}/{\bf 14.9}/{\bf 10.2}$ \\
\bottomrule
\end{tabular}
\end{table}

% \begin{table}
%   \caption{$n$-gram precision scores for the different models on the two datasets}
%   \label{table:ngram}
%   \centering
% \begin{tabular}{ccccc}
% \toprule
%  &\multicolumn{2}{c}{\bf \amazon}
% &
% \multicolumn{2}{c}{\bf \aol} 
% \\\cmidrule(r){2-3} \cmidrule(l){4-5}
%  & {\it Parameters} & {\it 1/2/3/4-gram}  & {\it Parameters} & {\it 1/2/3/4-gram}      \\
% \midrule
% {\bf Seq2Seq Attn.} & $112366657 (449.47MB)$ & $35.9/22.1/13.6/9.1$ & $105358732 (421.43MB)$ &  $28.9/13.9/9.7/8.3$ \\
% {\bf H-LSTM} & $125670465 (502.68MB)$ & $35.4/20.2/12.3/8.1$ & $118662540 (474.65MB)$ & $26.5/10.6/6.3/4.5$ \\
% {\bf M-NSRF} & $125670465 (502.68MB)$ & ${\bf 36.0}/20.7/12.9/8.6$ & $118662540 (474.65MB)$ & $27.0/11.0/6.7/4.9$\\
% {\bf Transformer} & $115485669 (461.94MB)$ & $34.0/20.6/12.6/7.6$  &   $104814396 (419.26MB)$  &$34.2/19.6/13.1/8.9$ \\
% {\bf MTN (Ours)}  & $116297701 (465.19MB)$ & $35.1/{\bf 26.4}/{\bf 19.0}/{\bf 13.1}$ & $101684540 (406.74MB)$  & ${\bf 35.7}/{\bf 21.5}/{\bf 14.9}/{\bf 10.2}$ \\
% \bottomrule
% \end{tabular}
% \end{table}

\subsection{Evaluation metrics}
We evaluated the performance of different models in terms of their $n$-gram precision scores, as done previously for  query suggestion by \cite{ACW2018}, and the cumulative BLEU scores \cite{PRWZ2002}. The $n$-gram scores are computed by counting the number of $n$-gram matches between the suggested or {\em candidate} queries, and the corresponding actual next or {\em reference} queries issued by the user. For instance, $1$-gram or unigram score is computed by comparing the individual tokens, while the $2$-gram or bigram score evaluates word pairs. These comparisons are made independent of the positions, i.e., without taking the order of tokens into account. However, the counting of matches is modified, based on actual frequency of tokens in the reference query, to ensure candidate queries are not overly rewarded for several occurrences of a matching word. We report these $n$-gram precision scores for $n \in \{1, 2, 3, 4\}$ to be consistent with the standard practice.  The BLEU score, additionally, imposes a brevity penalty on very short candidate queries. The score is known to correlate well with human judgements \cite{PRWZ2002}.  

\subsection{Results}
\begin{figure*}
\hskip -1.8cm
\begin{floatrow}
\ffigbox{%
  %\centering
  \includegraphics[width=6cm]{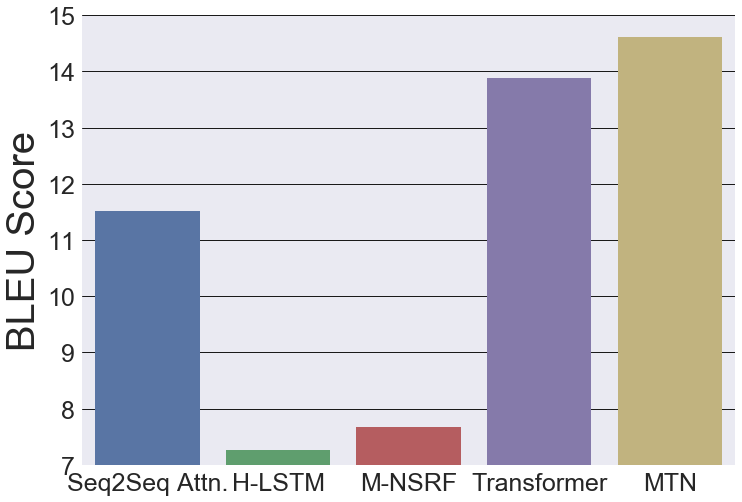}%
}{%
  \caption{BLEU scores on \aol test data \label{fig:aol_bleu}}%
}
\hskip -0.8cm
\capbtabbox{%
  \begin{tabular}{lcccr}
\toprule
{\bf Model} & {\bf Enc. Layers} & {\bf Dec. Layers} & {\bf BLEU} \\
\toprule \toprule
Transformer & (4Q, 0) & 4 & 13.90 \\
\midrule
Transformer & (5Q, 0) & 5 &  13.89 \\
\midrule
MTN (Ours) & (3Q, 2S) & 3 & 14.62 \\
\bottomrule
\end{tabular}

}{%
  \caption{(\aol data) Effect of different types of layers \label{table:aol_bleu}}%
}
\end{floatrow}
\end{figure*}

Table \ref{table:ngram} shows the model size (under single-precision floating-point representation), and $n$-gram precision scores for the different models on the two datasets. We indicate the best performing model in bold. We observe that M-NSRF performs better than the other methods in terms of $1$-gram score on the \amazon data, with Seq2Seq being a close second. However, note that almost all models perform reasonably well, and the gap between them is rather small. In contrast, MTN significantly outperforms all the other methods on the rest of the precision scores. Specifically, the discrepancy in performance of MTN relative to the next best algorithm, i.e. Seq2Seq with attention, is massive in each case: about $20\%$ ($2$-gram), and $40\%$ on  $3$-gram and $4$-gram. This clearly underscores that MTN is able to exploit the multiresolution structure much better than the rest. Similarly, as Table \ref{table:ngram} shows,  MTN registered remarkably higher precision scores than the baselines on \aol (note the model sizes for all methods are comparable). In fact, compared to the best recurrent model,  MTN scored 20\% higher on 1-gram and 4-gram, and 50\% higher on other precision scores. Fig. \ref{fig:aol_bleu} compares the BLEU score of the different methods on \aol data corresponding to the models from Table \ref{table:ngram}. We first observe that the fully attentional models (Transformer, MTN) outperform the recurrent models (Seq2Seq, H-LSTM, M-NSRF). We further observe that MTN obtained a much higher BLEU score than  Transformer (over $5\%$ improvement) and the best recurrent model (over $25\%$ improvement). Our results illustrate the benefits of employing MTNs for tasks with hierarchical structure.

We now provide more evidence that MTN teases out the hierarchical structure more effectively than Transformer.
Specifically, we show that session (i.e., level 2) layers  in MTNs cannot be supplanted by additional Transformer encoder layers without risking a substantial decrease in performance. We denote the query level encoder layers by $Q$ and session level encoder layers by $S$. Note that our MTN model (from Table \ref{table:ngram}) with 3 query layers, 2 session layers, and 3 decoder layers outperformed the optimized Transformer architecture with the same total number of layers (i.e. 8) split between encoder and decoder. As the Table \ref{table:ngram} shows, the performance of MTN could not be matched even by increasing the size of encoder and decoder stacks in the standard Transformer architecture. 

Finally, Table \ref{table:suggestions} shows a sample of queries suggested by MTN on \amazon. We also provide a sample of suggestions on the \aol data in the supplementary material. We found that MTN was often able to suggest new queries that reflected the intent in successive user searches over a short span. 
% % Note use of \abovespace and \belowspace to get reasonable spacing
% % above and below tabular lines.
% \begin{table}[t]
% \caption{Details of Data sets}
% \label{sample-table}
% \vskip 0.15in
% \begin{center}
% \begin{small}
% \begin{sc}
% \begin{tabular}{lcccr}
% \toprule
% Data set  & Type  & Sessions & Query Pairs\\
% \midrule
% & {\bf Training} & $4092254$ & $10964531$ \\
% \amazon & {\bf Validation} & $84111$ & $225775$ \\
% & {\bf Test} & $42349$ & $113579$ \\
% \bottomrule
% & {\bf Training} & $1628433$ & $4285507$ \\
% \aol & {\bf Validation} & $90932$ & $238890$ \\
% & {\bf Test} & $90507$ & $238289$ \\
% \bottomrule
% \end{tabular}
% \end{sc}
% \end{small}
% \end{center}
% \vskip -0.1in
% \end{table}

%\blindtext

% \begin{table}[t]
% \caption{Performance of different models on \amazon data}
% \label{sample-table}
% %\vskip 0.15in
% \begin{center}
% \begin{small}
% \begin{sc}
% \begin{tabular}{lcccr}
% \toprule
% {\bf Model} & $n$-gram scores ($n = 1/2/3/4$) & {\bf Parameters} \\
% \midrule
% Seq2seq w/ Attn. & $35.9/22.1/13.6/9.1$ & $112366657$ \\
% HLSTM & $35.4/20.2/12.3/8.1$ &  $125670465$ \\
% MNSRF & ${\bf 36.0}/20.7/12.9/8.6$ & $125670465$\\
% Transformer & $34.0/20.6/12.6/7.6$ & 
% $115485669$\\
% MTN (Ours)    & $35.1/{\bf 26.4}/{\bf 19.0}/{\bf 13.1}$ & $116297701$ \\
% \bottomrule
% \end{tabular}
% \end{sc}
% \end{small}
% \end{center}
% \vskip -0.1in
% \end{table}

% \begin{table}
%   \caption{Sample table title}
%   \label{sample-table}
%   \centering
%   \begin{tabular}{lll}
%     \toprule
%     \multicolumn{2}{c}{\amazon}                   \\
%     \cmidrule(r){1-3}
%     Name     & Description     & Size ($\mu$m)  \\
%     \midrule
%     Dendrite & Input terminal  & $\sim$100  \\
%     Axon     & Output terminal & $\sim$10    \\
%     Soma     & Cell body       & up to $10^6$  \\
%     \bottomrule
%   \end{tabular}
% \end{table}

\begin{table}
\setlength{\tabcolsep}{4pt}
\caption{Examples of query suggestions by MTN on \amazon data}
\label{table:suggestions}
%\centering
\begin{tabular}{l|l|l}
\toprule
{\bf Previous session queries} &  {\bf Predicted next query} &
{\bf User next query} \\
\midrule
\MakeLowercase{mini glad containers}, mini storage && \\containers, 
lunch bag cold & lunch bag cold pack & freezable lunch bag \\
\hline
lawn games for kids, lawn games, fun && \\ birthday games, games legged race & kids toys & water balloons \\
\hline
% motherhood maternity leggings, maternity & & \\ leggings,  maternity activewear shorts, & unicorn backpack&\\ unicorn backpack &  for girls & local honey\\
% \hline
 bore brush, drive wire brush, wire brush, &&\\ shank & metric wrench set & hex shank\\
 \hline 
cub cadet wheel bearings, cub cadet wheel &&\\ bushings, cub cadet wheel spacers, cub &cub cadet&\\ cadet hub &  hub assembly & cub cadet\\
\hline
scull bong, silicone bong, bubblers for &mini bong&\\ smoking weed, mini bong &  for smoking & mini hookah\\
\hline
vanity mirror with lights, vanity mirror, &makeup mirror&\\ makeup mirror &  with lights & mirror with lights\\
\hline
moana favors, moana plates and cups, &moana&\\
brown napkins paper, moana napkins paper &  party supplies & moana tag <unk> water\\
% \hline
%  iphone x case,  iphone x accessories, & wireless charger &\\
%   iphone x card holder, wireless charger &   iphone plus &  fast wireless charger\\
 \hline
%  prime pantry, swiffer wet mopping refills, &laundry detergent&\\ tilex mold and mildew, laundry detergent &  prime pantry  & paper towels\\
nightmare chess, lords of waterdeep board,&&\\ and games rising sun, azul game & lego batman &   fire table \\
\bottomrule
\end{tabular}
\end{table}

% \begin{table}[t]
% \caption{Performance of different models on \aol data}
% \label{sample-table}
% \begin{center}
% \begin{small}
% \begin{sc}
% \begin{tabular}{lcccr}
% \toprule
% Model & $1/2/3/4$-gram & #Parameters \\
% \midrule
% Seq2seq Attn. & $28.9/13.9/9.7/8.3$ & $105358732$ \\
% HLSTM & $26.5/10.6/6.3/4.5$ &  $118662540$ \\
% MNSRF & $27.0/11.0/6.7/4.9$ & $118662540$\\
% Transformer  & $34.2/19.6/13.1/8.9$ &  $104814396$\\
% MTN (Ours)  & ${\bf 35.7}/{\bf 21.5}/{\bf 14.9}/{\bf 10.2}$ & $101684540$ \\
% \bottomrule
% \end{tabular}
% \end{sc}
% \end{small}
% \end{center}
% \vskip -0.1in
% \end{table}

% \begin{figure}
%   \centering
%   \includegraphics[width=5cm]{Bleu_Score.png}
%   %\fbox{\rule[-.5cm]{0cm}{4cm} %\rule[-.5cm]{4cm}{0cm}}
%   %\caption{Sample figure caption.}
% \end{figure}

% {\bf  BLEU Scores $11.51, 7.26, 7.68, 13.89, 14.62$ }

%% file: conclusion.tex
\section{Conclusions} \label{conclusion}
We introduced multiresolution models that rely entirely on attention. Our models demonstrated strong empirical performance on two datasets pertaining to query recommendations. It would be interesting to use our framework for other tasks with hierarchical structure such as logical inference \cite{TBM2018}, where the recurrent models were found to perform better than the Transformer model. 

Our formalism paves way for interesting directions such as reducing the memory footprint of Transformer based models similar in spirit to the methods for compressing recurrent nets \cite{KSBKJV2018, ZWLW2018}. Such models could be deployed, e.g., on mobile phones and as conversational AI programs (chatbots).  

%% file: supplementary.tex
\section{Supplementary Material} \label{supplementary}
We now provide proofs for all the results stated in the main text.\\

\noindent \textbf{Proof of Proposition \ref{Prop1}}
\begin{proof}
Recall the standard equations for RNN from \eqref{RNN1_eq1}
\begin{eqnarray*}
h_t & = & \phi_h(x_t W_h + h_{t-1}U_h) \label{RNNeq1} \\
y_t & = & \phi_y(h_t W_y)~, \nonumber
\end{eqnarray*}
where $x_t$, $h_t$, and $y_t$ as row vectors; $W_h, U_h, W_y$ are weight matrices; and $\phi_h, \phi_y$ are activation functions (e.g. ReLU). We define a row vector 
$\tilde{x}^t = \text{Concat}(x_t, h_{t-1})$ that concatenates $x_t$ and $h_{t-1}$, and thus encapsulates both the recurrent state $h_{t-1}$ and the input $x_t$ to RNN at time $t$. We also form a matrix $\tilde{W}_h$ by stacking rows of $W_h$ atop $U_h$. We can do this since for the sum in \eqref{RNNeq1} to be well-defined, $W_h$ and $U_h$ must have the same number of columns. Thus, we can write   
\begin{eqnarray*}
h_t & = & \phi_h(\tilde{x}_t \tilde{W}_h) \\
y_t & = & \phi_y(h_t W_y)~.
\end{eqnarray*}
We collect all the $h_t$ together, and form a matrix $H$ that has $h_t$ as its row $t$. Likewise, we form $\tilde{X}$ by stacking $\tilde{x}_t$, and $Y$ by stacking $y_t$ as rows. 
Thus, extending the use of activations $\phi_h$ and $\phi_y$ from vectors to matrices,  we can write
\begin{eqnarray*}
H & = & \phi_h(\tilde{X} W_h)\\
Y & = & \phi_y(H W_y)~.
\end{eqnarray*}

Recall that in an RNN, at any time $t$, the only input information available is $\{\tilde{x}_1, \ldots, \tilde{x}_t\}$. Therefore, in order to trace the evolution of the RNN output, we mask the subsequent time steps by introducing a binary matrix $M_t$ that has $t$ rows, and same number of columns as the rows in the input matrix $X$. Specifically, we set $M_t(i, i) = 1$, and $M_t(i, j) = 0$ for $j \neq i$. In other words, $M_t$ is a {\em selection matrix} obtained by restricting an identity matrix to first $t$ rows. Then, defining $H_t$ as the matrix obtained by stacking the first $t$ rows of $H$, and likewise for $Y_t$, we can write
\begin{eqnarray*}
H_t & = & M_t H ~=~ M_t \phi_h(\tilde{X} W_h)\\
Y_t & = & \phi_y(H_t W_y)~, 
\end{eqnarray*}
which immediately yields
$$Y_t ~=~ \phi_y(M_t \phi_h(\tilde{X} W_h) W_y)~.$$
\end{proof}

\noindent \textbf{Proof of Proposition \ref{Prop2}}
\begin{proof}
We start with a transformer model with single-head attention. The extension to multi-head models is then straightforward. We reproduce the Transformer architecture from \cite{V2017} in Fig. \ref{fig:transformer}. 

Consider a transformer model with $L$ layers. For each layer $\ell \in [L]$, let $W_{\ell, 1} \in \mathbb{R}^{d \times d_f}, W_{\ell, 2} \in \mathbb{R}^{d_f \times d}$ be the parameters to be learned.  Let $\sigma(A)$ denote the probabilities obtained by applying softmax on each row of matrix $A$ independently. Let $\mathbb{I}_{n}$ denote the $n \times n$ identity matrix, and $1_n\in \mathbb{R}^n$ denote an $n$-dimensional column vector of all ones.  

We denote layer normalization by $\psi$, and softmax by $\sigma$. Then, we can write the single-head dot-product attention, pertaining to matrix $E$, scaled over dimension $d$ as 
$${\rm Att}_{SH}(E, d) ~~=~~ \sigma\left(\dfrac{EE^{\top}}{\sqrt{d}}\right)E~~.$$

The first sublayer in each layer composes layer normalization with the sum of attention and the input to the sublayer. Thus, the output of a sublayer on its input $E$ may be expressed as 
 $$O_1(E, d) =~ \psi(E + {\rm Att}_{SH}(E, d)) ~= \psi \left(E +  \sigma\left(\dfrac{EE^{\top}}{\sqrt{d}}\right)E \right) ~=~ \psi \left(\left(\mathbb{I}_n +  \sigma\left(\dfrac{EE^{\top}}{\sqrt{d}}\right) \right) E \right)~.$$ 
 The second sublayer transforms $O_1(E, d)$ via a feedforward network, adds $O_1(E,d)$ via a residual connection, and finally performs layer normalization. Omitting the bias terms for simplicity, we can express the effect of feedforward network with weights $W_1$ and $W_2$ on $O_1(E, d)$ as
 $${\rm FFN}(O_1(E, d)) = \phi(O_1(E,d) W_1)W_2~,$$
 where $\phi$ denotes the ReLU  activation. 
Thus, we obtain the following output from this sublayer:
\begin{eqnarray*} O_2(O_1(E, d)) & = & \psi \left( O_1(E, d) +  {\rm FFN}(O_1(E, d)) \right) \\
& = & \psi \left( O_1(E, d) +  \phi(O_1(E,d) W_1)W_2 \right)~. 
\end{eqnarray*}
Thus, we can view each encoder layer in the Transformer architecture as taking input $E$, and applying the composition $O_2 \circ O_1$. That is, we can write the output of an encoder layer as
$$O(E, d) ~=~ O_2(O_1(E, d)) = \psi \left( O_1(E, d) +  \phi(O_1(E,d) W_1)W_2 \right)~,$$
where $W_1$ and $W_2$ are weights specific to the layer. Since the output of each sublayer is a matrix, we can simplify the notation and replace the functional form of the outputs by equivalent matrices. Therefore, we have the following equations for the single head attention encoder that takes representation matrix $X \in \mathbb{R}^{n \times d}$ as the input (we assume the positional embeddings have already been added to initial embeddings to obtain $X$).  \\ 
% the \psi is the layer normalization, \sigma denotes the soft-max, I_n + \sigma (\cdot) denote the residual connection

%\newpage 
\begin{figure}
    \centering
    \includegraphics[width=8cm]{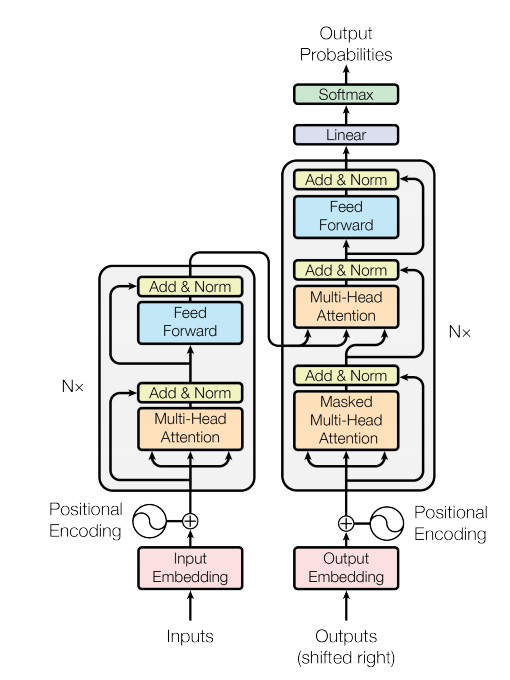}
    \caption{The Transformer architecture (source: \cite{V2017})
    \label{fig:transformer}}
\end{figure}

\noindent {\bf Single-head attention encoder}
\begin{eqnarray*}
E_{0} & = & X   \\
\hat{E}_{\ell} & = & \psi \left(\left(\mathbb{I}_n + \sigma\left(\dfrac{E_{\ell-1} E_{\ell-1}^{\top}}{\sqrt{d}}\right) \right)E_{\ell-1}\right), \quad \ell \in [L] \\
E_{\ell} & = &   \psi \left(\hat{E}_{\ell} +  \phi \left(\hat{E}_{\ell} W_{\ell, 1} \right) W_{\ell, 2}\right), \hspace*{1.7cm} \ell \in [L]~ \\
Y_{\ell} & = & E_{\ell}~, \hspace*{5.4cm} \ell \in [L]~. \\
\end{eqnarray*}
We now proceed to the multi-head attention encoder.

\noindent {\bf Multi-head attention encoder}\\
In a multi-head attention encoder, each of the $P$ heads works on a separate subspace of the embeddings. The attention at any layer $\ell$ is computed for each head $i \in [P]$ separately via projection matrices $Z_{\ell, i}^1$, $Z_{\ell, i}^2$ and $Z_{\ell, i}^3$, and these attentions are combined together via weights $Z_{\ell}^4$. Then, proceeding along the same lines as in the single-head setting, we can express the multi-head encoder as follows.
\begin{eqnarray*}
E_{0} & = & X  \\
\hat{E}_{\ell, i} & = & \left(\mathbb{I}_n + \sigma\left(\dfrac{E_{\ell-1} Z_{\ell, i}^1 Z_{\ell, i}^{2^{\top}} E_{\ell-1}^{\top}}{\sqrt{d}}\right) \right)E_{\ell-1} Z_{\ell, i}^3, \qquad \ell \in [L], ~~i \in [P]  \\
\hat{E}_{\ell} & = &  \psi \left(\text{Concat}(  \hat{E}_{\ell, 1}, \ldots, \hat{E}_{\ell, P})Z_{\ell}^4\right), \hspace*{2.4cm} \ell \in [L]\\
E_{\ell} & = &   \psi \left(\hat{E}_{\ell} +  \phi\left(\hat{E}_{\ell} W_{\ell, 1} \right) W_{\ell, 2}\right), \hspace*{2.9cm} \ell \in [L]~\\
Y_{\ell} & = & E_{\ell}~, \hspace*{6.6cm} \ell \in [L]~. \\
\end{eqnarray*}
The proposition follows by defining $Z_{\ell} = \{Z_{\ell}^4\} \cup \{(Z_{\ell, i}^1, Z_{\ell, i}^2, Z_{\ell, i}^3) : i \in [P]\}$, and  $\tilde{X}_{\ell} ~=~ \hat{E}_{\ell}$~.\\ 
\end{proof}

\noindent \textbf{Proof of Proposition \ref{Prop3}}
\begin{proof}
The decoder in a Transformer model (Fig. \ref{fig:transformer}) is laid out as a stack of layers. Each layer consists of three sublayers. The first sublayer employs {\em masked} attention on its input to prevent the flow of information from subsequent target tokens, and so preserve the auto-regressive property. We can implement this mask operation in the following way. 
Let $R$ be a matrix that has entries 1 everywhere on its diagonal and below (i.e. the lower triangular matrix), and $-\infty$ everywhere else. Let $\odot$ denote the Hadamard product, i.e., elementwise matrix mulplications. Then, we can write the single-head mask attention on input $D$ as
$${\rm Att}_{MSH}(D, d) ~~=~~ \sigma\left(R \odot \dfrac{DD^{\top}}{\sqrt{d}}\right)D~~.$$
The masked sublayer composes layer normalization with the sum of attention and the input to the sublayer. Thus, we may express the output of a masked single-head attention sublayer on input $D$ as 
 $$O_{M1}(D, d) =~ \psi(D + {\rm Att}_{MSH}(E, d)) ~= \psi \left(D +  \sigma\left(R \odot \dfrac{DD^{\top}}{\sqrt{d}}\right)D \right)~.$$ 
 
The second single-head attention sublayer generates attention by computing affinity between the output $O_{M1}(D, d)$ from the masked sublayer, and the output $Y_{enc}$ from the top of the encoder stack. Then it carries out an addition of this attention with $O_{M1}(D, d)$ via a residual connection, followed by layer normalization. Thus, we can express the output of this sublayer as
 $$O_{M2}(O_{M1}(D, d), Y_{enc}, d) = \psi \left(O_{M1}(D, d) + \sigma\left(\dfrac{O_{M1}(D, d)Y_{enc}^{\top}}{\sqrt{d}}\right)Y_{enc}\right)~.$$ 
 
 Finally, the third sublayer implements a feedforward transformation on $O_{M2}(O_{M1}(D, d), Y_{enc}, d)$ in an identical way to the second sublayer in each layer on the encoder. Thus, we can write the following equations for each layer $\ell \in [L]$ in a single-head attention decoder that receives $X'$ pertaining to the target tokens, and $Y_{enc}$ pertaining to the encoder output.  Note that we assume row $j$ in $X' \in \mathbb{R}^{m \times d}$  contains the position adjusted representations for token at position $j$ in the (partially decoded) target.
%  We prove the result for decoder with single-head attention. We omit the extension from single head to multi-head, since it is straightforward and follows along the lines of Proposition 2. Let $R$ be a matrix that has entries 1 everywhere on its diagonal and below (i.e. the lower triangular matrix), and $-\infty$ everywhere else. $R$ prevents the information flow due to tokens not yet decoded, and so preserves the auto-regressive property.  Let $\odot$ denote the Hadamard product, i.e., elementwise matrix mulplications. We have the following set of equations.
%  It is important to note that each of the matrices on the left side of the equations evolves with time since decoding is autoregressive, and thus only first $t$ rows are valid at time $t$ in each case. 
 
\noindent {\bf Single-head attention decoder}
\begin{eqnarray*}
D_{0} & = & X'  \\
\hat{D}_{\ell} & = & \psi\left(D_{\ell-1} + \sigma\left(R \odot \dfrac{D_{\ell-1} D_{\ell-1}^{\top}}{\sqrt{d}}\right) D_{\ell-1}\right)~, \hspace{3cm} \ell \in [L] \\
\tilde{D}_{\ell} & = & \psi\left(\hat{D}_{\ell} + \sigma\left(\dfrac{\hat{D}_{\ell} Y_{enc}^{\top}}{\sqrt{d}}\right) Y_{enc}\right)~, \hspace{4.7cm} \ell \in [L] \\
D_{\ell} & = &  \psi\left(\tilde{D}_{\ell} +  \phi\left(\tilde{D}_{\ell} W_{\ell, 1} \right) W_{\ell, 2}\right)~, \hspace{5cm} \ell \in [L]  \\
Y_{\ell} & = & D_{\ell}~, \hspace{8.8cm} \ell \in [L]~.
\end{eqnarray*}

Note that only first $t$ rows of $Y_{\ell}$ valid are valid at time $t$). The extension from single head to multi-head is straightforward and follows along the lines of Proposition \ref{Prop2}. We describe the decoder with multi-head attention below. \\

\noindent {\bf Multi-head attention decoder}
\begin{eqnarray*}
D_{0} & = & X'  \\
\hat{D}_{\ell, i} & = &  \left(\mathbb{I}_n + \sigma\left(R \odot \dfrac{D_{\ell-1} \hat{Z}_{\ell, i}^1 \hat{Z}_{\ell, i}^{2^{\top}} D_{\ell-1}^{\top}}{\sqrt{d}}\right) \right)D_{\ell-1} \hat{Z}_{\ell, i}^3, \qquad \ell \in [L], ~~i \in [P]  \\  
\hat{D}_{\ell} & = &  \psi \left(\text{Concat}(  \hat{D}_{\ell, 1}, \ldots, \hat{D}_{\ell, P})\hat{Z}_{\ell}^4\right), \hspace*{3.1cm} \ell \in [L]\\
\tilde{D}_{\ell, i} & = & \left(\mathbb{I}_n + \sigma\left(\dfrac{\hat{D}_{\ell}\tilde{Z}_{\ell, i}^1 \tilde{Z}_{\ell, i}^{2^{\top}} Y_{enc}^{\top}}{\sqrt{d}}\right) \right) Y_{enc} \tilde{Z}^3_{\ell, i}~,  \hspace*{1.8cm}  \ell \in [L], ~~i \in [P] \\ 
\tilde{D}_{\ell} & = &  \psi \left(\text{Concat}(  \tilde{D}_{\ell, 1}, \ldots, \tilde{D}_{\ell, P})\tilde{Z}_{\ell}^4\right), \hspace*{3.1cm} \ell \in [L]\\
D_{\ell} & = &  \psi\left(\tilde{D}_{\ell} +  \phi\left(\tilde{D}_{\ell} W_{\ell, 1} \right) W_{\ell, 2}\right)~, \hspace*{3.5cm} \ell \in [L]\\
Y_{\ell} & = & D_{\ell}~,  \hspace*{7.3cm}\ell \in [L]~.
\end{eqnarray*}
Note that the output evolves with time since decoding is autoregressive, and thus only first $t$ rows   $Y_{\ell, 1}, \ldots, Y_{\ell, t}$ of $Y_{\ell}$ are valid in the last equation above. Therefore, in order to trace the evolution of outputs $Y_{\ell}$ with time, we define a binary selection matrix $M_t$ for each time $t$ consisting of $t$ rows and $m$ columns (recall $m$ is number of rows in $X'$). Each row $r \in [t]$ of $M_t$ contains 1 at column $r$ and 0 elsewhere.  Then, since layer normalization $\psi$ operates on each row independently, we can express the decoder outputs for layer $\ell$ up to time $t$ as
$$Y_{\ell, t} =  \psi\left(M_t \tilde{D}_{\ell} +  M_t \phi\left(\tilde{D}_{\ell} W_{\ell, 1} \right) W_{\ell, 2}\right)~,$$
where we note that $\tilde{D}_{\ell}$ depends on both $X'$ and $Y_{enc}$. The proposition follows immediately when we define the following collection of multi-head weights for $\ell \in [L]$: $$Z_{\ell} = \{\hat{Z}_{\ell}^4, \tilde{Z}_{\ell}^4\} \cup \{(\hat{Z}_{\ell, i}^1, \hat{Z}_{\ell, i}^2, \hat{Z}_{\ell, i}^3, \tilde{Z}_{\ell, i}^1, \tilde{Z}_{\ell, i}^2, \tilde{Z}_{\ell, i}^3) : i \in [P]\}~.$$  \\
\end{proof}
\newpage

\noindent \textbf{Proof of Proposition \ref{Prop4}}
\begin{proof}
We now sketch the evolution of an MTN encoder. We will focus on single-head attention since it conveys the essential ideas. The extension to multi-head attention is straightforward, and follows along the lines of Propositions 
\ref{Prop2} and \ref{Prop3}, and thus omitted. 

Let $Q_S = \{q_1, \ldots, q_{|S|}\}$ be the queries in a given session $S$, where we denote the number of queries in $S$ by $|S|$. Without loss of generality,\footnote{As is common practice, if the queries are of variable length, we can pad the queries to ensure they all have the same number of tokens as the longest query.} let each query $q \in Q_S$ consist of $n$ tokens. We indicate the embeddings pertaining to query $q$ by appropriate subscripts, e.g., $X_q \in \mathbb{R}^{n \times d}$ denotes the input token representations for $q$. Let $L_1$ be the number of layers in the query level encoder, and $L_2$ in the session level encoder of MTN. We use notation $E_{1, \ell, q}$ to denote the output embeddings for query $q$ at layer $\ell \in [L_1]$ of query level encoder. 
% We denote the output embeddings for query $q$ at layer $\ell \in [L_2]$ of the session encoder by $y_{2, \ell, q}$ to indicate that the output embedding for each query is a vector.
Moreover, we denote the weights for layer $\ell$ at level $r \in \{1, 2\}$ by $W_{r, \ell,1}$ etc.
Since the query encoding component of an MTN encoder is the same as a Transformer encoder, we can reproduce the expressions from Proposition \ref{Prop2} for dynamics  
at the query level.      

\noindent {\bf Single-head attention query level encoder}
\begin{eqnarray*}
E_{1, 0, q} & = & X_q~,  \hspace{8.5cm} q \in Q_S \\
\hat{E}_{1, \ell, q} & = & \psi \left(\left(\mathbb{I}_n + \sigma\left(\dfrac{E_{1, \ell-1, q} E_{1, \ell-1, q}^{\top}}{\sqrt{d}}\right) \right)E_{1, \ell-1, q}\right), \hspace*{0.8cm} \ell \in [L_1]~, q \in Q_S  \\
{E}_{1, \ell, q} & = &   \psi \left(\hat{E}_{1, \ell, q} +  \phi \left(\hat{E}_{1, \ell, q} W_{1, \ell, 1} \right) W_{1, \ell, 2}\right), \hspace*{2.1cm} \ell \in [L_1]~, q \in Q_S \\
Y_{1, \ell, q} & = & E_{1, \ell, q}~,\hspace*{6.6cm} \ell \in [L_1]~, q \in Q_S \\
\tilde{y}_{1, L_1, q} & = &
w_{proj} Y_{1, L_1, q}~, \hspace{7cm} q \in Q_S~.
\end{eqnarray*}

Note the additional equation at the end.
% $\tilde{Y}_{1, L_1, q} \in \mathbb{R}^{n \times d}$ for $q \in Q_S$. 
MTN projects $Y_{1, L_1, q}$ via an $n$-dimensional row vector $w_{proj}$ to obtain $\tilde{y}_{1, L_1, q} \in \mathbb{R}^{1 \times d}$.  The query embeddings $\tilde{Y}_{1, L_1} \triangleq \{\tilde{y}_{1, L_1, q} : q \in Q_S\} \in \mathbb{R}^{|S| \times d}$ are adjusted for position and feed into the masked session level. To avoid extra notation, we add the query position encodings to $\tilde{Y}_{1, L_1}$, and call the resulting embeddings $\tilde{Y}_{1, L_1}$ as well. Let $R_S$ be a $|S| \times |S|$ matrix that has entries 1 everywhere on its diagonal and below (i.e. the lower trinagular matrix) and $-\infty$ everywhere else. Let $\odot$ denote the Hadamard product, i.e., elementwise matrix multiplications. Let $e_{q} \in \{0,1\}^{1 \times |S|}$ be a row vector with 1 at position $j \in [|S|]$ if $q$ is the $j^{th}$ query in $Q_S$, and 0 at all other positions. We adapt the expressions from Proposition \ref{Prop3} to sketch the evolution of the output at the session level of MTN. \\

\noindent {\bf Single-head attention session level encoder}
\begin{eqnarray*}
E_{2, 0} & = & \tilde{Y}_{1, L_1}  \\
\hat{E}_{2, \ell} & = & \psi\left(E_{2, \ell-1} + \sigma\left(R_S \odot \dfrac{E_{2, \ell-1} E_{2, \ell-1}^{\top}}{\sqrt{d}}\right) E_{2, \ell-1}\right)~, \hspace{0.8cm} \ell \in [L_2] \\
E_{2, \ell} & = &  \psi\left(\hat{E}_{2, \ell} +  \phi\left(\hat{E}_{2, \ell} W_{2, \ell, 1} \right) W_{2, \ell, 2}\right)~, \qquad \hspace{2.2cm} \ell \in [L_2] \\
Y_{2, \ell} & = & E_{2, \ell}~,  \qquad \hspace{6.6cm} \ell \in [L_2] \\
y_{2, \ell, q} & = & e_q E_{2, \ell}~, \qquad \hspace{6.3cm} \ell \in [L_2]~, q \in Q_S ~.\\
% Y_{2, \ell, q} & = & E_{2, \ell, q}~, \qquad \hspace{6.4cm} \ell \in [L_2] \\
% \tilde{y}_{2, L_2, q}  & = & \psi(y_{2, L_2, q} ~+~ \tilde{y}_{1, L_1, q})~, \hspace{6.3cm}  q \in Q_S  
\end{eqnarray*}
Note that the last equation extracts out the embedding vector pertaining to query $q$. This vector is added to each row of the embedding matrix $Y_{1, L_1, q}$ defined under the query level encoder, and layer normalization is performed. As a result, the correlations of $q$ with the queries preceding $q$ in the session are accounted for in the individual token embeddings. Let $\hat{Y}_{2, L_2, q} \in \mathbb{R}^{n \times d}$ be formed by stacking $n$ copies of $y_{2, \ell, q}$. Thus, the output of an MTN encoder can be expressed as 
$$\tilde{Y}_{2, L_2, q} ~=~ \psi(Y_{1, L_1, q} ~+~ \hat{Y}_{2, L_2, q})~.$$
Note that the position of query $q$ in the session serves as a time index $t(q) \in [|S|]$ for $q \in Q_S$. Thus, we can view the evolution of the output of MTN encoder for $q$ and all queries preceding $q$ via $t(q)$. Specifically, let $\tilde{Y}_{2, L_2} \in \mathbb{R}^{n|S| \times d}$ be formed by stacking matrices $\tilde{Y}_{2, L_2, q_1}, \ldots, \tilde{Y}_{2, L_2, q_{|S|}}$  vertically. Likewise, we define matrices $Y_{1, L_1}$ and $\hat{Y}_{2, L_2}$. We define a binary selection matrix $M_{t} \in \mathbb{R}^{t \times n|S|}$ as in Proposition $\ref{Prop3}$, i.e., each row $r \in [t]$ of $M_t$ contains 1 at column $r$ and 0 in all the other columns.  Then, since at time $t(q)$ pertaining to position of $q$, only the first $n t(q)$ rows of $\tilde{Y}_{2, L_2}$ are valid, we can write the evolution of outputs for all queries up to time $t(q)$as 
$$\tilde{Y}_{2, L_2, t(q)} = \psi\left(M_{n t(q)}(Y_{1, L_1} ~+~ \hat{Y}_{2, L_2})\right)~.$$
The proposition follows by noting that 
we can write $\hat{Y}_{2, L_2, q}$ may be written as $C_{t(q)} Y_{2, L}$, where $C_{t(q)} \in \{0, 1\}^{n \times |S|}$ contains 1 at each entry in column $t(q)$, and 0 everywhere else.  
\end{proof}

% and 
% Since no subsequent   

% Here, the last equation indicates the effect of layer normalization on the sum of (1) output embeddings following the projection of embeddings yielded by query level, and (2) output embeddings for each query produced by the session encoder.  Note that at any time $t \in [|S|]$, where $t$ is an index over the queries in the session, no subsequent embeddings are valid. Therefore, stacking the embeddings 
% \text{(Only first $q$ rows of $\hat{D}_{2, \ell}$ \text{ and } $Y_{2, \ell}$ are valid when $q$ is processed}

\section*{Additional experimental results}
We now show some sample query suggestions produced by MTN on \aol in Table \ref{table:aol_suggestions}. 

\begin{table}[h]
\setlength{\tabcolsep}{4pt}
\caption{Examples of query suggestions by MTN on \aol}
\label{table:aol_suggestions}
%\centering
\begin{tabular}{l|l|l}
\toprule
{\bf Previous session queries} &  {\bf Predicted next query} &
{\bf User next query} \\
\midrule
spanish dictionary, homework help, & spanish english&\\ spanish english <unk> &  translation &
spanish english translator\\
\hline
summer camps for year olds in &&\\ wilmington nc,
summer camps in & jelly beans summer & jelly beans family \\wilmington nc, jelly beans summer camp & camp in new york & skating center  \\
\hline
driving directions, travelocity, &&\\ driving directions & mapquest & tyler perry\\ 
\hline
www myspace, myspace, www myspace & www myspace com & www myspace com\\
\hline
l l bean, road runner sports, men nylon pants & men clothing & men nylon wind pants \\
\hline
all the road running, cd stores, best buy & circuit city & fye\\
\hline
easy make ahead food, make ahead & best potato salad &make ahead no cook\\ memorial day meals, best potato salad &  recipe &  desserts\\
\hline
orbitz, northwest airlines, orbitz & expedia & northwest airlines\\
\hline
coldwell banker, thyroid disease &thyroid disease&\\ alcoholism, thyroid disease &  symptoms &  thyroid\\
\hline
bed and breakfast in st augustine, &&\\ brunswick georgia, golden isles resorts & golden retriever resort & simon island \\
\hline
www mysprint com, sprint, telephone & telephone numbers & telephone numbers\\
\hline
usa today com, cnn com, bartleby com & free encyclopedia & free encyclopedia\\
\hline
busta rhymes, bow wow & lil wayne & fresh azimiz\\
\hline
standford university, havard,& university of& \\ havard university &  phoenix  & yale university\\
\hline
nyse eslr, nyse hl, amex bgo & amex bema gold & nyse hl\\
\hline
shears, styling shears, hair styling shears & hair styles & hair cutting techniques\\
\hline
spirit airlines, orlando airlines, usa & delta airlines & orlando airlines\\
\hline
university of phoenix diploma, copy of &&\\ university of phoenix diploma, the best & the best on line&on line\\ on line fully <unk> university &   colleges & pharmacy degree \\
\hline
pastel braided rugs, craigs list, craigslist & ebay & craigslist washington state\\
\hline
macys, ralph lauren home, pottery barn & crate and barrell & tommy <unk>\\
\bottomrule
\end{tabular}
\vskip -0.1in
\end{table}